\documentclass[11pt]{article}
\usepackage{coling2018}
\usepackage{times}
\usepackage{url}
\usepackage{latexsym}

\usepackage{amsmath}
\usepackage{amssymb}

\usepackage{multirow}
\usepackage[ruled,vlined,commentsnumbered,longend]{algorithm2e}
\usepackage{xcolor}

\usepackage{mathrsfs}
\usepackage{graphicx}
\usepackage{subfig}
\usepackage{amsfonts}
\usepackage{booktabs}
\usepackage{float}
\usepackage{subfig}
\usepackage{tikz-dependency}
\usepackage{pgfplots}

\pgfplotsset{compat=1.14}

\usepackage{bm}
\usepackage{url}

\usepackage{mathtools}

\newcommand{\norm}[1]{\left\lVert#1\right\rVert}

\def\x{\mathbf{x}}

\def\bv{\mathbf{v}}

\def\h{\textbf{h}}
\def\e{\mathbf{e}}

\def\p{\mathbf{p}}

\def\bh{\mathbf{h}}

\def\s{\mathbf{s}}
\def\m{\mathbf{m}}
\def\v{\mathbf{v}}
\def\bh{\mathbf{h}}

\DeclareMathOperator*{\softmax}{softmax}
\def\tran{^\mathrm{\scriptscriptstyle T}}

\SetKwComment{Comment}{/*}{*/}

\title{Information Aggregation  via Dynamic Routing for Sequence Encoding}

\author{Jingjing Gong, Xipeng Qiu\thanks{\hspace{1mm} Corresponding Author}, Shaojing Wang, Xuanjing Huang\\
 Shanghai Key Laboratory of Intelligent Information Processing, Fudan University\\
School of Computer Science, Fudan University\\
\{jjgong15, xpqiu, sjwang17, xjhuang\}@fudan.edu.cn\\
}
\date{}

\begin{document}
\maketitle
\begin{abstract}
While much progress has been made in how to encode a text sequence into a sequence of vectors, less attention has been paid to how to aggregate these preceding vectors (outputs of RNN/CNN) into fixed-size encoding vector. Usually, a simple max or average pooling is used, which is a bottom-up and passive way of aggregation and lack of guidance by task information.
In this paper, we propose an aggregation mechanism to obtain a fixed-size encoding with a dynamic routing policy. The dynamic routing policy is dynamically deciding that \textit{what} and \textit{how much} information need be transferred from each word to the final encoding of the text sequence. Following the work of Capsule Network, we design two dynamic routing policies to aggregate the outputs of RNN/CNN encoding layer into a final encoding vector. Compared to the other aggregation methods, dynamic routing can refine the messages according to the state of final encoding vector. Experimental results on five text classification tasks show that our method outperforms other aggregating models by a significant margin. Related source code is released on our github page\footnote{https://github.com/FudanNLP/Capsule4TextClassification}.
\end{abstract}

\section{Introduction}
\label{intro}

Learning the distributed representation of text sequences, such as sentences or documents, is crucial to a wide range of important natural language processing applications. A primary challenge is how to encode the variable-length text sequence into a fixed-size vector, which should fully capture the semantics of text.

Many successful text encoding methods usually contain three key steps: (1) converting each word in a text sequence into its embedding; (2)
taking as input the sequence of word embeddings, and computing the context-aware representation for each word with a recurrent neural network (RNN) ~\cite{hochreiter1997long,chung2014empirical} or convolutional neural network (CNN)~\cite{collobert2011natural,kim2014convolutional};
(3) summarizing the sentence meaning into a fixed-size vector by an aggregation operation.
Then, these models are trained by combining a downstream task in a supervised or unsupervised way.

Currently, much attention is paid to the first two steps, while the aggregation step is less emphasized on.
Some simple aggregation methods, such as max (or average) pooling, is used to sum the RNN hidden states or convolved vectors, computed in the previous step, into a single vector. This kind of methods aggregate information in a bottom-up and passive way and are lack of the guide of task information.
Recently, several works employ self-attention mechanism ~\cite{lin2017structured,yang2016hierarchical} on top of the recurrent or convolutional encoding layer to replace simple pooling. A basic assumption is that the words (or even sentences) are not equally important. One or several task-specific context vectors are used to assign a different weight to each word and
select task-specific encodings. The context
vectors are parameters learned jointly with other parameters during the training process. These attentive aggregation can select task-dependent information. However, the context vectors are fixed once learned.

%These aggregation methods summarize information of sentences in different ways. Despite succussing in many NLP tasks,

In this paper, we regard the aggregation as a routing problem of how to deliver the messages from source nodes to target nodes. In our setting, the source nodes are the outputs of a recurrent or convolutional encoding layer, and the target nodes are one or several fixed-size encoding vectors to represent the meaning of the text sequence.

From this viewpoint, both the pooling and attentive aggregations are a \textit{fixed} routing policy without considering the state of the final encoding vectors. For example, the final encoding vectors could receive some redundancy information from different words. The fixed routing policy cannot avoid this issue. Therefore, we wish for a new way to aggregate information according to the state of the final encoding.

In  recent promising work of capsule network~\cite{DBLP:conf/nips/SabourFH17}, a dynamic routing policy is proposed and proven to be more effective than the max-pooling routing. Inspired by their idea, we introduce a text sequence encoding model with dynamic routing mechanism. Specifically, we propose two kinds of dynamic routing policies. One is the standard dynamic routing policy same as the capsule network, in which the source node decides what and how many messages are sent to different target nodes. The other is the reversed dynamic routing policy, in which the target node decides what and how many messages may be received from different source nodes.

Experimental results on five text classification tasks show that the dynamic routing policy outperforms other aggregation methods, such as max pooling, average pooling, and self-attention by a significant margin.

\section{Background: general sequence encoding for text classification}

% As shown in Figure \ref{fig:dyroute}, our overall encoding model of text sequence consists of three layer: (1) embedding layer, (2) recurrent encoding layer, (3) aggregating layer.
%
% After aggregating process, we have fixed-size encoding vector, which is further fed to an MLP for the final prediction in text classification.

In this section, we are going to introduce a general text classification framework. It consists of an Embedding Layer, Encoding Layer, Aggregation Layer and Prediction Layer.

\subsection{Embedding Layer}
Given a text sequence with words $S=w_1,w_2,\cdots,w_L$. Since the words are symbols that could not be processed directly using prominent neural architectures, so we first map each word into a $d$ dimensional embedding vector,
\begin{align}
	X = [\x_1,\x_1,\x_2,\cdots,\x_L].
\end{align}

In order to transfer knowledge from a vast unlabeled corpus, the embeddings can be taken from the pre-trained word embedding, such as Glove ~\cite{pennington2014glove}.

\subsection{Encoding Layer}
However, each word representation in $X$ is still independent with each other. To gain some dependency between adjacent words, we then build a bi-directional LSTM (BiLSTM) layer ~\cite{hochreiter1997long} to incorporate forward and backward context information of a sequence. Then we can get phrase-level encoding $\bh_t$ of a word by concatenating forward $\bh_t^f$ and backward output vector $\bh_t^b$ correspond to the target word.

\begin{align}
	\bh_t^f &= \text{LSTM}(\bh^f_{t-1}, \x_t),\\
	\bh_t^b &= \text{LSTM}(\bh^b_{t+1}, \x_t),\\
	\bh_t &= [\bh_t^f; \bh_t^b].
\end{align}

Thus, the outputs of BiLSTM encoder are a sequence of vectors
\begin{align}
	H=[\bh_1,\bh_2,\cdots,\bh_L].
\label{eq:hidden}
\end{align}

\subsection{Aggregation Layer}

Encoding layer only models dependency between adjacent words, but the final prediction of the text requires a fix-length vector. Therefore we need aggregate information from variable length sequence to a single fix-length vector. There are several different ways of aggregation such as max or average pooling, and context-attention.
%and a method proposed in this paper --- Dynamic Routing which will be introduced in detail in Section \ref{sec:dyrout}.

\paragraph{Max or Average Pooling}
Max or Average pooling is a simple way of aggregating information, which does not require extra parameters and is computationally efficient~\cite{kim2014convolutional,zhao2015self,lin2017structured}.
In the process of modeling natural language, max or average pooling is performed along the time dimension.
\begin{align}	
\e^{max}&=\max([\bh_1,\bh_2,\cdots,\bh_L]),\\
\e^{avg}&=\frac{1}{L} \sum_{i=1}^L {\mathbf{h}_i},
\end{align}
For example,in Equation 6 the $\max$ operation is performed on each dimension of $\mathbf{h}$ along time dimension. And in Equation 7 the average operation is  performed along time dimension.

Max pooling is empirically better at aggregating long sentences than average pooling. We assume it's because that, the actual word that contributes to the classification problem is far less than the number of words that contain in a long sentence. Information from important words is weakened by a large population of ``boring" words. %Even for shorter sentences, only few words are important to the objective of problem in hand.

\paragraph{Self-Attention}
 As has been stated previously, average pooling is prone to weaken important words when the sentence is longer. Self-Attention assigns each word a weight to indicate the importance of a word depending on the task on hand. A few words that are crucial to the task will be emphasized while the ``boring'' words are ignored. The self-attention process is formulated as follows:
\begin{align}	
	u_i&=\mathbf{q}^T\mathbf{h}_i, \\
	a_i&=\frac{\exp(u_{i})}{\sum_k{\exp(u_{k})}}\label{eq:softmax}, \\
	\e^{attn}&=\sum_{i=1}^L {a_i \cdot \mathbf{h}_i}\label{eq:weighted_sum}
\end{align}
First, we need a task-specific trainable query $\mathbf{q} \in \mathbb{R}^{d}$ to calculate similarity weight between query and each contextually encoded word. Then the corresponding weights are normalized across time dimension using softmax normalization function Eq. \ref{eq:softmax}, after that the aggregated vector is simply a weighted sum of the input sequence in Eq. \ref{eq:weighted_sum}.

% The aggregation layer can alleviate the burden of encoding layer. Each LSTM hidden state is only expected to provide shorter term context information around each word, while the higher level semantics, which requires longer term dependency, can be picked up directly by the aggregation layer.

\subsection{Prediction Layer}

Then we feed the encoding $\e$ to the input of a multi-layer perceptron (MLP), followed by a softmax classifier.
 \begin{equation*}
	 \p(\cdot|\e) = \softmax(\mathrm{MLP}(\e))
 \end{equation*}
 where $\p(\cdot|\e)$ is the predicted distribution of different classes given the representation vector $\m$.

\section{Aggregation via Dynamic Routing}
\label{sec:dyrout}
In this section, we will formally introduce dynamic routing in detail.
The goal of dynamic routing is to encode the meaning of $X$ into $M$ fix-length vectors
\begin{align}
	V=[\bv_1,\bv_2,\cdots,\bv_M].
\end{align}

To transfer information from a variable number of representation $H$ to a fixed number of vectors $V$,
a key problem we need to slove is to properly design a routing policy of information transfer. In other words, \textit{what} and \textit{how much} information is to be transferred from $\bh_i$ to $\bv_j$.

%A much simple but effective way is max or average pooling, and a relatively complicated way is weighted averaging by self-attention~\cite{lin2017structured}. However, the notion of summing up elements in the attention mechanism is still very primitive, and it can be something more complex.
Although self-attention has been applied in aggregation, the notion of summing up elements in the attention mechanism is still very primitive.
Inspired by the capsule networks \cite{DBLP:conf/nips/SabourFH17}, we propose a dynamic routing aggregation (DR-AGG) mechanism to compute the final encoding of text sequence.

Following the definition of capsule networks, we call each encoding vector, or a group of neurons, as a capsule. Thus, $H$ denotes the input capsules, and $V$ denotes the output capsules.

A message vector $\m_{i\rightarrow j}$ denotes the information to be transferred from $\bh_i$ to $\bv_j$.
\begin{equation}
	\m_{i\rightarrow j} = c_{ij} f(\bh_i,\theta_j),
\end{equation}
where $c_{ij}$ indicates proportionally how much information is to be transferred, and $f(\bh_i,\theta_j)$ is a one-layer fully-connected network parameterized by $\theta_j$, indicating which aspect of information is to be transferred.

The output capsule $\v_j$ first aggregates all the incoming messages
\begin{equation} \label{eq:sss}
	\s_j = \sum_{i=1}^L \m_{i\rightarrow j},
\end{equation}
and then squashes $\s_j$ to confine $|\s_j|\in (0,1)$ to a probability,
\begin{equation}
	\label{eq:squash}
	\v_j = \frac{\norm{\s_j}^2}{1+\norm{\s_j}^2} \frac{\s_j}{\norm{\s_j}}
\end{equation}

%We want the length of the output vector of a capsule to represent the probability that the entity represented by the capsule is present in the current input. We therefore use a non-linear "squashing" function to ensure that short vectors get shrunk to almost zero length and long vectors get shrunk to a length slightly below 1. We leave it to discriminative learning to make good use of this non-linearity

\begin{figure}[t]
 \centering
 \subfloat[Aggregation via Dynamic Routing]{\includegraphics[width=0.6\textwidth,height=20em]{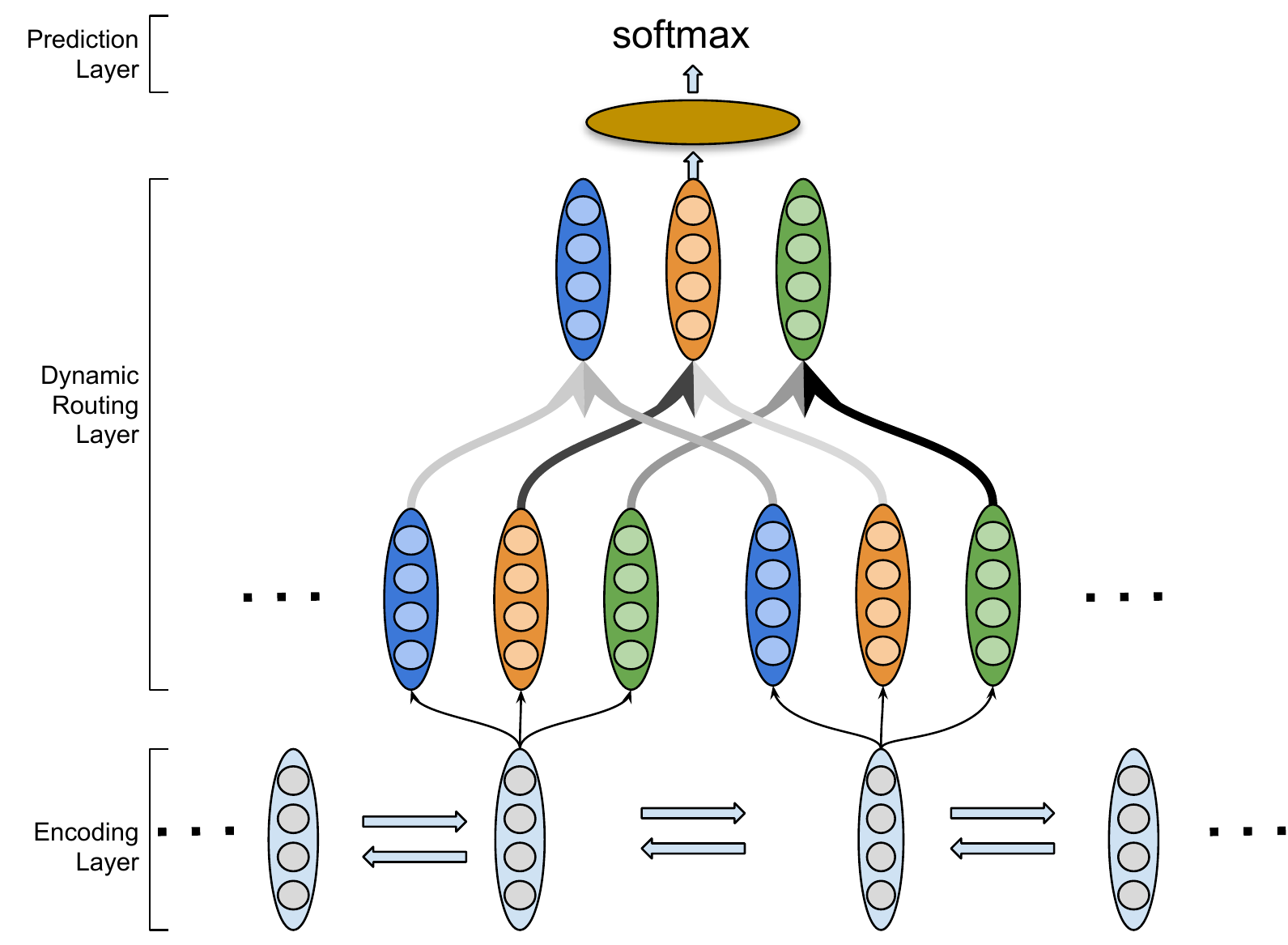}}
 \hspace{2em}
  \subfloat[Detialed dynamic routing process]{\includegraphics[width=0.3\textwidth,height=13em]{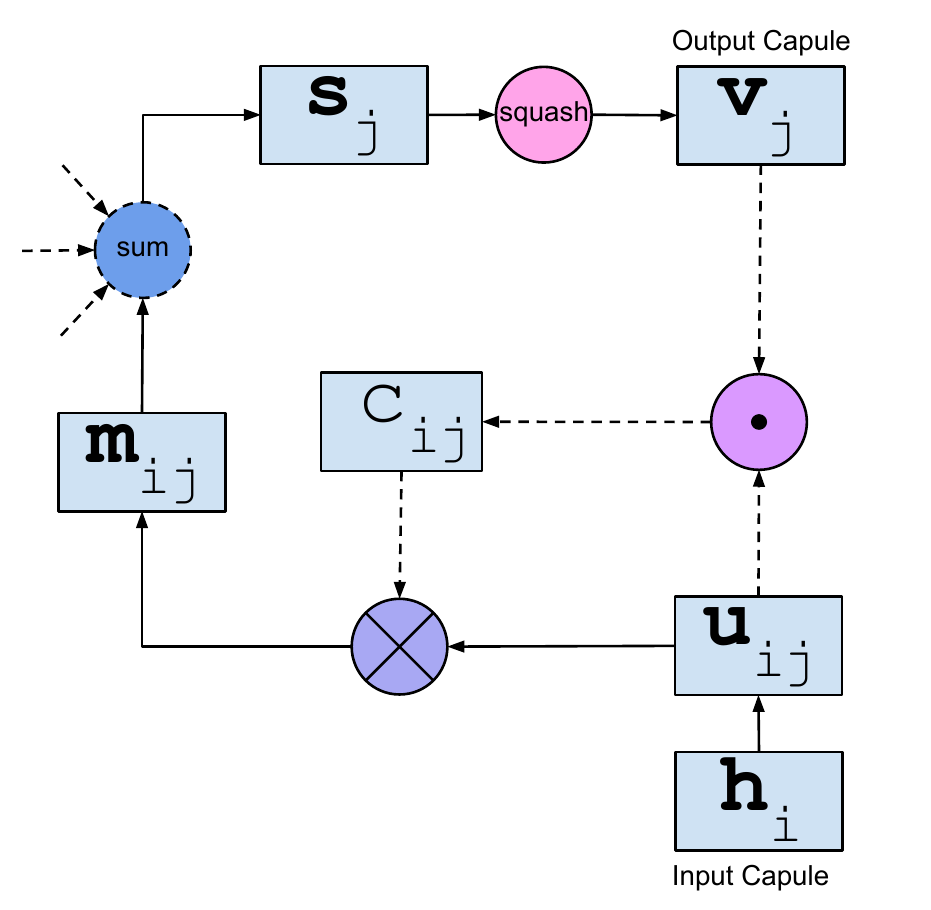}}
  \caption{Diagram of dynamic routing for sequence encoding. (a) is the overall dynamic routing diagram, the width of edges between capsules  indicate the amount of information transfered, which is refined iteratively. (b) is a detailed iterative process of transferring information from capsule $\h_i$ to capsule $\v_j$, where $\odot$ is a inner product operation and $\otimes$ is a element-wise product.}\label{fig:dyroute}
\end{figure}

\paragraph{Dynamic Routing Process}
The dynamic routing process is implemented by an iterative process of refining the coupling coefficient $c_{ij}$, which define proportionally how much information is to be transferred from $\bh_i$ to $\bv_j$.

The coupling coefficient $c_{ij}$ is computed by
\begin{align}
	c_{ij} &= \frac{\exp(b_{ij})}{\sum_k{\exp(b_{ik})}},\label{eq:rout_soft}\\
b_{ij} &\leftarrow b_{ij} + \v_j\tran f(\bh_i,\theta_j), \label{eq:rout_soft-b}
\end{align}
where $b_{ij}$ is the log probabilities, initialized with 0.

The coefficients $c_{ij}$ is computed using a softmax function, and $\sum_{j=1}^M c_{ij}=1$. Therefore, the total amount of information transferred from capsule $h_i$ is proportionally summed to one.

When an output capsule $\v_j$ receives the incoming messages, its state will be updated and the coefficient $c_{ij}$ is also re-computed for each input capsule. Thus, we iteratively refine the route of information passing, towards an instance dependent and context aware encoding of a sequence.
After the text sequence is encoded into $M$ capsules,
We  map these capsules into vector representation by simply concatenating all capsules,
 \begin{equation}
	 \e= [\v_1; \ldots; \v_M].
 \end{equation}

Figure~\ref{fig:dyroute} gives an illustration of the dynamic routing mechanism. The detailed dynamic routing algorithm is further described in detail in Algorithm \ref{algo:dyrout}.

\begin{algorithm}%[th]
\SetAlgoLined
\KwData{Input Capsules: $\bh_1,\bh_2,\cdots,\bh_L$, Maximum number of Iterations: $T$}
\KwResult{Output Capsules: $\bv_1,\bv_2,\cdots,\bv_M$ }
 Initialize $b_{ij} \leftarrow 0$ \;
 \For{$t=1$ \KwTo $T$}{
   Compute the routing coefficients $c_{ij}$ for all $i\in[1,L],j\in[1,M]$ \tcc*[r]{Eq.\ref{eq:rout_soft}}

  Update all the output capsule $\v_j,j\in[1,M]$ \tcc*[r]{Eq. \ref{eq:sss} and \ref{eq:squash}}

  Update $b_{ij}$ for all $i\in[1,L],j\in[1,M]$
  \tcc*[r]{Eq. \ref{eq:rout_soft-b}}

 }
 \KwRet{$\bv_1,\bv_2,\cdots,\bv_M$}
 \caption{Dynamic Routing Algorithm}
	\label{algo:dyrout}
\end{algorithm}

\paragraph{Reversed Dynamic Routing Process}

In standard DR-AGG, an input capsule decides what proportion of information can be transferred to an output capsule. We also explore a reversed dynamic routing, in which the output capsule decides what proportion of information should be received from an input capsule. The only difference between reversed dynamic routing and standard dynamic routing is how the softmax function was applied to the log probabilities $[b_{ij}]_{L\times M}$. Instead of normalizing each row of $[b_{ij}]_{L\times M}$ as is done in standard DR-AGG, reverse dynamic routing normalizes each column of $[b_{ij}]_{L\times M}$,
\begin{equation}
	\label{eq:Rrout_soft}
	c_{ij} = \frac{\exp(b_{ij})}{\sum_k{\exp(b_{kj})}}
\end{equation}

Other detail of reversed dynamic routing is the same as the standard dynamic routing.

%The reversed DR-AGG works like the multi-hop memory network~\cite{sukhbaatar2015end,kumar2015ask}, to iterative aggregate information. 
The reversed DR-AGG works like the multi-hop memory network in iteratively aggregating information~\cite{sukhbaatar2015end,kumar2015ask}. 

\subsection{Analysis}

The DR-AGG is somewhat like attention mechanism~\cite{bahdanau2014neural,DBLP:conf/nips/VaswaniSPUJGKP17}.however, there are differences.

In standard DR-AGG, each input capsule (encoding of each word) is employed as query vector to assign a proportion weight to each output capsule, and then sends messages to the output capsules in proportion. Thus, for all input capsules the total amount of messages sent from an input capsule are the same.

In reversed DR-AGG, each output capsule is used as query vector to assign a proportion weight to each input capsule and then receives messages from the input capsules in proportion. Thus, for all output capsules the total amount of message received by an output capsule is same.

The major difference between DR-AGG and self-attention \cite{lin2017structured,yang2016hierarchical} is that the query vector of self-attention is task dependent trainable parameters learned during the training phase, while the query vector of DR-AGG is each input or output capsule which is instance dependent and dynamically updated.

Additionally, the self-attention aggregation collects information in a bottom-up way, without considering the state of the final encoding. It is hard to avoid the problems of information redundancy and information loss.
While in the standard DR-AGG, each word can iteratively decide \textit{what} and \textit{how much} information is to be sent to the final encoding.

\section{Hierarchical Dynamic Routing for Long Text}

The dynamic routing mechanism can aggregate the text sequence with any length, therefore it is able to handle long texts directly, such as the whole paragraphs or documents.

To further enhance the efficiency and scalability of information aggregation, we adopt a hierarchical dynamic routing mechanism to handle the long text. The hierarchical routing strategy can exploit more parallelization and speed up training and inference process. A similar strategy is also used in~\cite{yang2016hierarchical}.

Concretely, we split a document into sentences, and apply the proposed dynamic routing mechanism on word and sentence levels separately. We first encode each sentence into a fixed-length vector, then convert the sentence encodings into document encoding.

\section{Experiment}
We test the empirical performance of our proposed model on 5 benchmark datasets for document and sentence level classification and compare our proposed model to other competitor models.
\subsection{Datasets}
\begin{table}[t!] \setlength{\tabcolsep}{3pt}
  \centering%\small
  \begin{tabular}{cccccccc}
    \toprule
\textbf{Dataset} &\textbf{Type} &\textbf{Train Size} & \textbf{Dev. Size} &\textbf{Test size} & \textbf{Classes}  &\textbf{Averaged Length}   &\textbf{Vocabulary Size}\\
    \midrule % line
Yelp 2013         &Document      &62522       &7773    &8671     &5    &189 	&29.3k\\
Yelp 2014         &Document      &183019      &22745   &25399 	 &5    &197     &49.6k\\
IMDB          	  &Document      &67426      &8381   &9112 	     &10    &395    &61.1k\\
SST-1         	  &Sentence      &8544      &1101   &2201	     &5    &18      &16.3k\\
SST-2         	  &Sentence      &6920      &872   &1821	     &2    &19      &14.8k\\
    \bottomrule
  \end{tabular}
  \caption{Statistics of the five datasets used in this paper}
  \label{tab:dataset}
\end{table}

  %To evaluate the effectiveness of the capsule classification model, we experiment in the following five dataset. And the first three are document classification. So that the information need to be extract from any sentence’s any place and learn to select the right sentence and pay attention to that sentence’s exact section come to be more important.

  To evaluate the effectiveness of our proposed aggregation method, we have conducted experiments on 5 datasets, the statistics of experimented datasets are shown in Table \ref{tab:dataset}. As shown in the table, Yelp-2013, Yelp-2014, and IMDB are document level datasets, while SST-1 and SST-2 are sentence level datasets. Note that we use the same document level datasets provided in \cite{tang2015learning}.
   % of which the first three are document classifications.
  % So information need to be extract from arbitrary block of any sentences and it has to learn to select the right sentence which, then, we put emphasis on the exact section of it.\\

% \textbf{Yelp reviews} is a dataset of Yelp Dataset Challenge based on Yelp website. We evaluate our model on 2013 and 2014 Yelp Dataset Challenge. Each example is composed of several sentences and a review score.Each example is composed of several sentences and a review score.
\textbf{Yelp reviews} Yelp-2013 and Yelp-2014 are reviews from Yelp, each example consists of several review sentences and a rating score range from 1 to 5 (higher is better).

\textbf{IMDB} is a movie review dataset extracted from IMDB website. It is a multi-sentence dataset that for each example there are several review sentences. A rating score range from 1 to 10 is also associated with each example.
 % this dataset is built to get the scores of movie reviews extracted from the IMDB website. There usually are several sentences in one review write by audiences of the movie and along with the score.

\textbf{SST-1} Stanford Sentiment Treebank is a movie review dataset which has been parsed and further splited to train/dev/test set \cite{socher2013recursive}. For each example in the dataset, there exists only one sentence and a label associated with it. And the labels can be one of \{negative, somewhat negative, neutral, somewhat positive, positive\}.
%  Sentiment Treebank is labeled and parsed movie reviews, introduced in \cite{wang2016learning}, which is contain 11855 records. There are 5 labels in the dataset,
% they are , respectively, negative, somewhat negative, neutral, somewhat positive, positive.

\textbf{SST-2} This dataset is a binary-class version of SST-1, with neutral reviews removed and the remaining reviews categorized to either negative or positive.
 % but there are only two classes of labels.

\subsection{Training}

Given a training set $\{x^{(i)}, t^{(i)}\}_{i = 1}^N$, where $x^{(i)}$ is an example of the training set and $t^{(i)}$ is the corresponding label, the goal is to minimize the cross-entropy loss $\mathcal{J}(\theta)$:
\begin{equation}
\mathcal{J}(\theta) \!=\! -\frac{1}{N}\!\!\sum_i\! \log p(t^{(i)}|x^{(i)} ; \theta) \!+\! \lambda||\theta||^2_2,
\end{equation}
where $\theta$ represents all of the parameters.

The Adam optimizer is applied to update the parameters \cite{kingma2014adam}. Table \ref{tab:hypersetting} displays the detailed hyper-parameter settings. To prevent overfitting,  the L2 regularization term is introduced to our loss function.
 We also adopt early stop strategy, The training process will be stopped after seven epochs of no improvement on development set is observed.To further avoid overfitting, dropout is applied before the biLSTM encoder and hidden layer of classifier \textbf{MLP}.

The mini-batch size is set to 32 for document level dataset, 64 for sentence level dataset, examples are sampled from a sliding bucket to speed up the training process. Data is sorted by the length of sentence, and we first sample a window on the sorted data, we call the window ``sliding bucket" and then sample a batch of examples from the sliding bucket, we double the window size after an epoch of no improvement on development set, through such a strategy, we are able to considerably speed up training while retaining randomness. Also, batch size is halved after an epoch of no improvement on development set until it reaches the low bound batch size. We also utilize a data preparation queue to parallelize data preparation and training.

Word embedding is initialized from pre-trained Glove ~\cite{pennington2014glove}.
We randomly initialize word vectors for words that doesn't appear in Glove. Network weights are initialized with Xavier Normalization ~\cite{glorot2010understanding}. 
A more detailed hyper-parameter setting can be referred to hyper-parameter Table \ref{tab:hypersetting}. And hyper-parameters are determined using grid search strategy.

\begin{table}[t!] \setlength{\tabcolsep}{3pt}
  \centering%\small
  \begin{tabular}{lccccc}
    \toprule
  &\textbf{Yelp-2013} &\textbf{Yelp-2014} & \textbf{IMDB} &\textbf{SST-1} & \textbf{SST-2}\\
\midrule % line
Embedding size           		&300    	&300       	&300  		&300     &300     \\
LSTM hidden unit         		&200		&200       	&200    	&200     &200     \\
Capsule dimension	        	&200		&200       	&200    	&200     &200     \\
Capsule number	         		&5			&5       	&5    		&5     	 &5     \\
Iteration number	         	&3			&3       	&3    		&3     	 &3     \\
Regularization rate        		&1e-5      	&1e-5
&1e-6   	&1e-6    &1e-5    \\
Initial learning rate         	&0.0001     &0.0002     &0.0001   	&0.0001  &0.0003  \\
learning rate decay        		&0.9      	&0.9      	&0.95   	&0.95    &0.95    \\
learning rate decay steps       &1000      	&1000     	&1000   	&500     &500  \\
Initial Batch size         		&32      	&32     	&32   		&64      &64   \\
Batch size low bound         	&32      	&32     	&32   		&16      &16   \\
% Init sliding bucket size      &5000      	&5000     	&5000   	&5000    &5000 \\
Dropout rate         			&0.2      	&0.2      	&0.2   		&0.2     &0.5  \\

    \bottomrule
  \end{tabular}
  \caption{Detailed hyper-parameter settings}
  \label{tab:hypersetting}
\end{table}

\subsection{Experimental Results}
We evaluate several aggregation methods  on five text classification datasets, in which Yelp-2013, Yelp-2014 and IMDB are document-level datasets, and SST-1 and SST-2 are sentence-level datasets. Since max pooling, average pooling and self-attention are most related to our proposed DR-AGG, we mainly compare DR-AGG to these three methods. 

Table \ref{tab:results} gives the results for different methods, the last two rows are our model ( standard DR-AGG  and reversed DR-AGG), the table shows that our proposed dynamic routing performed the best on all datasets. 
In document-level text classification, specifically Yelp 2013 Yelp 2014 and IMDB, DR-AGG outperforms previous models’ best results by 2.5\%, 3.0\% and 1.6\% respectively. In sentence-level text classification, such as SST-1 SST-2, our model also achieves better results. 
Compared to max pooling, average pooling and self-attention, which are closely related to our model, DR-AGGs significantly improves the performance. For example the standard DR-AGG outperforms the max pooling approach by 1\%, 1.8\%, 4\%,2.5\% and 0.4\% on Yelp 2013,Yelp 2014, IMDB, SST-1 and SST-2. It empirically shows that our proposed dynamic routing policy is the most effective method on aggregating information. 

It is worth to note the reversed DR-AGG is inferior to the standard 
DR-AGG by a small margin, although it has also achieved better results than the other aggregation methods and SOTA approaches. 
As discussion before, the reversed DR-AGG have much resemblance with the  attention using output capsule as query vector. Not all of the input capsules would be selected by the reversed DR-AGG, while in the standard DR-AGG, the information of all the input capsules need be sent to the output capsules.

\begin{figure}[t]%\small
  \centering
  \pgfplotsset{width=0.5\textwidth}
  \begin{tikzpicture}
    \begin{axis}[
    xlabel={Iteration},
	ylabel={Accuracy on test set},
    legend entries={1 caps,2 caps,3 caps,4 caps},
    mark size=1.5pt,
    ymajorgrids=true,
    grid style=dashed,
    legend pos= north east,
    legend style={font=\footnotesize,line width=1.0pt,mark size=.5pt,
            /tikz/every even column/.append style={column sep=0.5em}},
            smooth,
    ]
%    \addplot [black,mark=square] table [x index=0, y index=8] {PM_length_analyse.txt};

    \addplot [blue,dashed,mark=square*] table [x index=0, y index=2] {data/caps1iter.txt};
	\addplot [red,dashed,mark=square*] table [x index=0, y index=2] {data/caps2iter.txt};
	\addplot [violet,dashed,mark=square*] table [x index=0, y index=2] {data/caps3iter.txt};
	\addplot [yellow,dashed,mark=square*] table [x index=0, y index=2] {data/caps4iter.txt};
    \end{axis}
\end{tikzpicture}
\caption{Relationship between test accuracy and routing iteration, where the vertical axis denotes test accuracy and the horizontal axis denotes routing iteration. When the iteration is set to 3 performance peaks on several different capsule number setting}
 % When identifying the iteration to update routing coefficients $c_{ij}$ ,we make some attempts that  we set the iteration round number from 1 to 5. And finding that when the iteration round number is 3,our model can give the best performance. That means during every batch data training, dynamic routing update the coefficients for choosing what and how much information to transfer. In contrast, self-attention only update the attention coefficients one time in every batch data iteration. Because routing section need a appropriate iteration round time to learn the adaptive routing coefficients and to select the truly important information.

\label{fig:capsiter}
\end{figure}
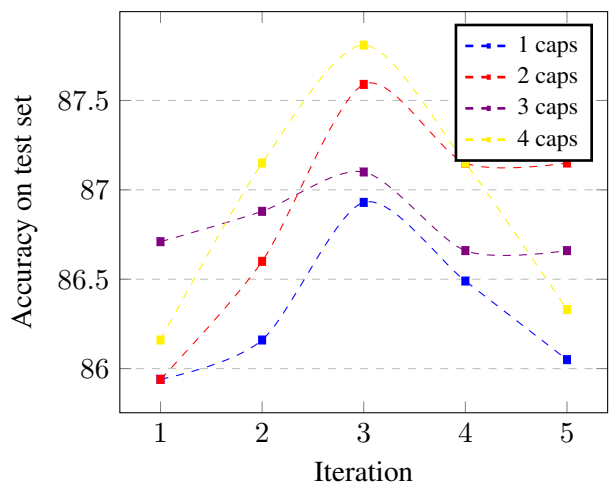

\paragraph{Effects of Iterative Routing} We also study how the iteration number affect the performance of aggregation on the SST-2 dataset.  Figure \ref{fig:capsiter} shows the comparison of 1 - 5 iterations in the standard DR-AGG. The capsule number is set to 1, 2, 3 and 4 for each comparison respectively. We found that  the performances on several different capsule number setting reach the best when iteration is set to 3. The results indicate the dynamic routing is contributing to improve the performance.

\paragraph{Visualization}
Additionally, we visualize how much information each input capsule sends to the output capsules. As shown in Table \ref{tab:weight_visual}, the visualization experiment was conducted with the setting on three output capsules. 
The $i$-th column represents the $i$-th input capsule, while the $j$-th row is the $j$-th output capsule. The color density of each word denotes the proportion $c_{ij}$ in equation \ref{eq:rout_soft}.  
A deeper color indicates more information of the concerned word is routed to the output capsule.

Intuitively, the different part of the sentence is routed to three different capsules. In another word, each capsule has a different perspective or focus of the sequence. Therefore, DR-AGG can avoid  the problem of information redundancy and information missing.

\label{exp}

\begin{table}[t!] %\setlength{\tabcolsep}{5pt}
  \centering%\small
  \begin{tabular}{lccccc}
    \toprule
&\textbf{Yelp-2013} &\textbf{Yelp-2014} &\textbf{IMDB} & \textbf{SST-1}  &\textbf{SST-2}  \\
    \midrule
RNTN+Recurrent~\cite{socher2013recursive} &57.4	&58.2	&40.0	&-	&-	\\
CNN-non-static~\cite{kim2014convolutional}&-		&-		&- 		&48.0	&87.2	 \\
Paragraph-Vec~\cite{le2014distributed}  	& -		& - 	& - 	& 48.7 & \textbf{87.8}   \\
MT-LSTM (F2S)~\cite{DBLP:conf/emnlp/LiuQCWH15} &- &- 	&- 		&49.1 	&87.2  \\
UPNN(np UP)~\cite{tang2015learning} 		&57.7	&58.5	&40.5	&- 	&-  \\
UPNN(full)~\cite{tang2015learning} 		& 59.6	&60.8 	& 43.5 &-  &-   \\
Cached LSTM~\cite{DBLP:journals/corr/XuCQH16}  &59.4  &59.2  &42.1  &- &-  \\
\midrule
Max pooling         	&61.1              &61.2              &41.1              &48.0               &87.0      \\
Average pooling         &60.7              &60.6              &39.1              &46.2               &85.2      \\
Self-attention       &61.0              &61.5 		 	   &43.3 			   &48.2		         &86.4		 \\

\midrule
Standard DR-AGG         &\textbf{62.1} &\textbf{63.0} &\textbf{45.1} &\textbf{50.5}  &87.6 \\
Reverse DR-AGG &61.6          &62.5          &44.5          &49.3 			 &87.2 \\
    \bottomrule
\end{tabular}
\caption{Experimental result comparison on five datasets. For the document-level datasets, hierarchical aggregation is used for both self-attention and DR-AGGs.}
\label{tab:results}
\end{table}

\begin{table}[t] \setlength{\tabcolsep}{10pt}
\begin{center}
\begin{tabular}{l}
\toprule
(1) \color{blue!17.11!white}{so} \color{blue!39.94!white}{relentlessly} \color{blue!21.96!white}{wholesome} \color{blue!28.90!white}{it} \color{blue!26.95!white}{made} \color{blue!28.46!white}{me} \color{blue!27.17!white}{want} \color{blue!13.36!white}{to} \color{blue!20.76!white}{swipe} \color{blue!10.81!white}{something} \color{blue!11.06!white}{.}\\
(2) \color{blue!51.46!white}{so} \color{blue!11.71!white}{relentlessly} \color{blue!14.31!white}{wholesome} \color{blue!33.77!white}{it} \color{blue!49.26!white}{made} \color{blue!54.32!white}{me} \color{blue!73.28!white}{want} \color{blue!94.33!white}{to} \color{blue!81.99!white}{swipe} \color{blue!98.31!white}{something} \color{blue!98.42!white}{.}\\
(3) \color{blue!51.43!white}{so} \color{blue!80.35!white}{relentlessly} \color{blue!83.73!white}{wholesome} \color{blue!57.32!white}{it} \color{blue!43.79!white}{made} \color{blue!37.22!white}{me} \color{blue!19.55!white}{want} \color{blue!12.31!white}{to} \color{blue!17.25!white}{swipe} \color{blue!10.88!white}{something} \color{blue!10.52!white}{.}\\
\bottomrule
\end{tabular}
\end{center}
\caption{A visualization to show the perspective of a sentence from 3 different upper level capsule. A deeper color indicates more information of the associated word is routed to the corresponding capsule.}
% After we set the iteration round number to 3, our model gives the best performance. To explain what and how much information has been transferred to the output vector, we visualize the weights of the three capsules on the input sentence. In the table, if the color of the word is heavier, the coefficient weight is bigger that means the routing mechanism pay more attention to those parts.}
\label{tab:weight_visual}
\end{table}

\section{Related Work}

Currently, much attention has been paid to how developing a sophisticated encoding models to capture the long and short term dependency information in a sequence. Specific to text classification task, most of the models cannot deal with the texts of several sentences (paragraphs, documents), such as MV-RNN \cite{socher2012semantic}, RNTN \cite{socher2013recursive}, CNN \cite{kim2014convolutional}, AdaSent \cite{zhao2015self}, and so on. The simple neural bag-of-words model can deal with long texts, but it loses the word order information. PV \cite{le2014distributed} works in an unsupervised way, and the learned vector cannot be fine-tuned on the specific task. There are also many works ~\cite{DBLP:conf/emnlp/LiuQCWH15,DBLP:journals/corr/XuCQH16,DBLP:conf/emnlp/0001DL16} to improve LSTM's ability to carrying information for a long distance.

A line of orthogonal researches ~\cite{lin2017structured,yang2016hierarchical,Shen2018DISAN,shen2018bi}
is to introduce attention mechanism~\cite{DBLP:conf/nips/VaswaniSPUJGKP17} to weighted average the outputs of CNN/RNN layer. The attention mechanism  can effectively reduce the burden of CNN/RNN. The CNN/RNN encoding layer is only expected to extract local context information for each word, while the global semantics of text sequence can be aggregated from the local encoding vectors.

The attention based aggregation collects information in a bottom-up way, without considering the state of the final encoding. It is hard to avoid the problems of information redundancy or information lost.
An improved idea is to use multi-hop attention, like memory network~\cite{sukhbaatar2015end,kumar2015ask}, to iterative aggregate information. This idea is equivalent to our proposed reversed dynamic routing mechanism.

Different from the attention based aggregation methods, aggregation via dynamic routing is iteratively deciding that \textit{what} and \textit{how much} information need be transfer to the final encoding of each word.
%The dynamic routing~\cite{sabour2017dynamic} is also a promising information aggregation approach.

\section{Conclusion}
In this paper, we focus on how to obtain a fixed-size encoding of text sequence by aggregating the encodings of each word. Although we use LSTM hidden states as word encoding in this paper, the other word encodings, such as convolved n-gram, could be alternatively used. We introduced a fixed-size encoding of text sequence with dynamic routing mechanism. Experimental results of five text classification tasks show that the model outperforms other encoding models by a significant margin.

In the future, we would like to investigate more sophisticated routing policy for better encoding the text sequence. Besides, dynamic routing should also be useful to improve the encoder in the sequence-to-sequence tasks~\cite{sutskever2014sequence}.

%As a downside of our proposed model, the current training method heavily relies on downstream applications, thus we are not able to train it in an unsupervised way. The major obstacle towards enabling unsupervised learning in this model is that during decoding, we don’t know as prior how the different rows in the embedding should be divided and reorganized. Exploring all those possible divisions by using a neural network could easily end up with overfitting. Although we can still do unsupervised learning on the proposed model by using a sequential decoder on top of the sentence embedding, it merits more to find some other structures as a decoder

\bibliographystyle{acl}
\bibliography{nlp}

\begin{thebibliography}{}

\bibitem[\protect\citename{{Bahdanau} \bgroup et al.\egroup
  }2014]{bahdanau2014neural}
D.~{Bahdanau}, K.~{Cho}, and Y.~{Bengio}.
\newblock 2014.
\newblock Neural machine translation by jointly learning to align and
  translate.
\newblock {\em ArXiv e-prints}, September.

\bibitem[\protect\citename{Cheng \bgroup et al.\egroup
  }2016]{DBLP:conf/emnlp/0001DL16}
Jianpeng Cheng, Li~Dong, and Mirella Lapata.
\newblock 2016.
\newblock Long short-term memory-networks for machine reading.
\newblock In {\em Proceedings of the 2016 Conference on Empirical Methods in
  Natural Language Processing, {EMNLP} 2016, Austin, Texas, USA, November 1-4,
  2016}, pages 551--561.

\bibitem[\protect\citename{Chung \bgroup et al.\egroup
  }2014]{chung2014empirical}
Junyoung Chung, Caglar Gulcehre, KyungHyun Cho, and Yoshua Bengio.
\newblock 2014.
\newblock Empirical evaluation of gated recurrent neural networks on sequence
  modeling.
\newblock {\em arXiv preprint arXiv:1412.3555}.

\bibitem[\protect\citename{Collobert \bgroup et al.\egroup
  }2011]{collobert2011natural}
Ronan Collobert, Jason Weston, L{\'e}on Bottou, Michael Karlen, Koray
  Kavukcuoglu, and Pavel Kuksa.
\newblock 2011.
\newblock Natural language processing (almost) from scratch.
\newblock {\em The Journal of Machine Learning Research}, 12:2493--2537.

\bibitem[\protect\citename{Glorot and Bengio}2010]{glorot2010understanding}
Xavier Glorot and Yoshua Bengio.
\newblock 2010.
\newblock Understanding the difficulty of training deep feedforward neural
  networks.
\newblock In {\em International conference on artificial intelligence and
  statistics}, pages 249--256.

\bibitem[\protect\citename{Hochreiter and Schmidhuber}1997]{hochreiter1997long}
Sepp Hochreiter and J{\"u}rgen Schmidhuber.
\newblock 1997.
\newblock Long short-term memory.
\newblock {\em Neural computation}, 9(8):1735--1780.

\bibitem[\protect\citename{Kim}2014]{kim2014convolutional}
Yoon Kim.
\newblock 2014.
\newblock Convolutional neural networks for sentence classification.
\newblock {\em arXiv preprint arXiv:1408.5882}.

\bibitem[\protect\citename{Kingma and Ba}2014]{kingma2014adam}
Diederik Kingma and Jimmy Ba.
\newblock 2014.
\newblock Adam: A method for stochastic optimization.
\newblock {\em arXiv preprint arXiv:1412.6980}.

\bibitem[\protect\citename{Kumar \bgroup et al.\egroup }2015]{kumar2015ask}
Ankit Kumar, Ozan Irsoy, Jonathan Su, James Bradbury, Robert English, Brian
  Pierce, Peter Ondruska, Ishaan Gulrajani, and Richard Socher.
\newblock 2015.
\newblock Ask me anything: Dynamic memory networks for natural language
  processing.
\newblock {\em arXiv preprint arXiv:1506.07285}.

\bibitem[\protect\citename{Le and Mikolov}2014]{le2014distributed}
Quoc~V. Le and Tomas Mikolov.
\newblock 2014.
\newblock Distributed representations of sentences and documents.
\newblock In {\em Proceedings of ICML}.

\bibitem[\protect\citename{Lin \bgroup et al.\egroup }2017]{lin2017structured}
Zhouhan Lin, Minwei Feng, Cicero Nogueira~dos Santos, Mo~Yu, Bing Xiang, Bowen
  Zhou, and Yoshua Bengio.
\newblock 2017.
\newblock A structured self-attentive sentence embedding.
\newblock {\em arXiv preprint arXiv:1703.03130}.

\bibitem[\protect\citename{Liu \bgroup et al.\egroup
  }2015]{DBLP:conf/emnlp/LiuQCWH15}
Pengfei Liu, Xipeng Qiu, Xinchi Chen, Shiyu Wu, and Xuanjing Huang.
\newblock 2015.
\newblock Multi-timescale long short-term memory neural network for modelling
  sentences and documents.
\newblock In {\em Proceedings of the 2015 Conference on Empirical Methods in
  Natural Language Processing, {EMNLP} 2015, Lisbon, Portugal, September 17-21,
  2015}, pages 2326--2335.

\bibitem[\protect\citename{Pennington \bgroup et al.\egroup
  }2014]{pennington2014glove}
Jeffrey Pennington, Richard Socher, and Christopher Manning.
\newblock 2014.
\newblock Glove: Global vectors for word representation.
\newblock In {\em Proceedings of the 2014 conference on empirical methods in
  natural language processing (EMNLP)}, pages 1532--1543.

\bibitem[\protect\citename{Sabour \bgroup et al.\egroup
  }2017]{DBLP:conf/nips/SabourFH17}
Sara Sabour, Nicholas Frosst, and Geoffrey~E. Hinton.
\newblock 2017.
\newblock Dynamic routing between capsules.
\newblock In {\em Advances in Neural Information Processing Systems 30: Annual
  Conference on Neural Information Processing Systems 2017, 4-9 December 2017,
  Long Beach, CA, {USA}}, pages 3859--3869.

\bibitem[\protect\citename{Shen \bgroup et al.\egroup }2018a]{Shen2018DISAN}
Tao Shen, Tianyi Zhou, Guodong Long, Jing Jiang, Shirui Pan, and Chengqi Zhang.
\newblock 2018a.
\newblock {DISAN}: Directional self-attention network for {RNN}/{CNN}-free
  language understanding.
\newblock In {\em AAAI Conference on Artificial Intelligence}.

\bibitem[\protect\citename{Shen \bgroup et al.\egroup }2018b]{shen2018bi}
Tao Shen, Tianyi Zhou, Guodong Long, Jing Jiang, and Chengqi Zhang.
\newblock 2018b.
\newblock Bi-directional block self-attention for fast and memory-efficient
  sequence modeling.

\bibitem[\protect\citename{Socher \bgroup et al.\egroup
  }2012]{socher2012semantic}
Richard Socher, Brody Huval, Christopher~D Manning, and Andrew~Y Ng.
\newblock 2012.
\newblock Semantic compositionality through recursive matrix-vector spaces.
\newblock In {\em Proceedings of EMNLP}, pages 1201--1211.

\bibitem[\protect\citename{Socher \bgroup et al.\egroup
  }2013]{socher2013recursive}
Richard Socher, Alex Perelygin, Jean~Y Wu, Jason Chuang, Christopher~D Manning,
  Andrew~Y Ng, and Christopher Potts.
\newblock 2013.
\newblock Recursive deep models for semantic compositionality over a sentiment
  treebank.
\newblock In {\em Proceedings of EMNLP}.

\bibitem[\protect\citename{Sukhbaatar \bgroup et al.\egroup
  }2015]{sukhbaatar2015end}
Sainbayar Sukhbaatar, Jason Weston, Rob Fergus, et~al.
\newblock 2015.
\newblock End-to-end memory networks.
\newblock In {\em Advances in Neural Information Processing Systems}, pages
  2431--2439.

\bibitem[\protect\citename{Sutskever \bgroup et al.\egroup
  }2014]{sutskever2014sequence}
Ilya Sutskever, Oriol Vinyals, and Quoc~VV Le.
\newblock 2014.
\newblock Sequence to sequence learning with neural networks.
\newblock In {\em Advances in Neural Information Processing Systems}, pages
  3104--3112.

\bibitem[\protect\citename{Tang \bgroup et al.\egroup }2015]{tang2015learning}
Duyu Tang, Bing Qin, and Ting Liu.
\newblock 2015.
\newblock Learning semantic representations of users and products for document
  level sentiment classification.
\newblock In {\em Proceedings of the 53rd Annual Meeting of the Association for
  Computational Linguistics and the 7th International Joint Conference on
  Natural Language Processing (Volume 1: Long Papers)}, volume~1, pages
  1014--1023.

\bibitem[\protect\citename{Vaswani \bgroup et al.\egroup
  }2017]{DBLP:conf/nips/VaswaniSPUJGKP17}
Ashish Vaswani, Noam Shazeer, Niki Parmar, Jakob Uszkoreit, Llion Jones,
  Aidan~N. Gomez, Lukasz Kaiser, and Illia Polosukhin.
\newblock 2017.
\newblock Attention is all you need.
\newblock In {\em Advances in Neural Information Processing Systems 30: Annual
  Conference on Neural Information Processing Systems 2017, 4-9 December 2017,
  Long Beach, CA, {USA}}, pages 6000--6010.

\bibitem[\protect\citename{Xu \bgroup et al.\egroup
  }2016]{DBLP:journals/corr/XuCQH16}
Jiacheng Xu, Danlu Chen, Xipeng Qiu, and Xuanjing Huang.
\newblock 2016.
\newblock Cached long short-term memory neural networks for document-level
  sentiment classification.
\newblock {\em CoRR}, abs/1610.04989.

\bibitem[\protect\citename{Yang \bgroup et al.\egroup
  }2016]{yang2016hierarchical}
Zichao Yang, Diyi Yang, Chris Dyer, Xiaodong He, Alex Smola, and Eduard Hovy.
\newblock 2016.
\newblock Hierarchical attention networks for document classification.
\newblock In {\em Proceedings of the 2016 Conference of the North American
  Chapter of the Association for Computational Linguistics: Human Language
  Technologies}, pages 1480--1489.

\bibitem[\protect\citename{Zhao \bgroup et al.\egroup }2015]{zhao2015self}
Han Zhao, Zhengdong Lu, and Pascal Poupart.
\newblock 2015.
\newblock Self-adaptive hierarchical sentence model.
\newblock {\em arXiv preprint arXiv:1504.05070}.

\end{thebibliography}
\end{document}